\documentclass{bmvc2k}

\title{Convolutional Simplex Projection Network for Weakly Supervised Semantic Segmentation}

\addauthor{Rania Briq}{briq@iai.uni-bonn.de}{1}
\addauthor{Michael Moeller}{Michael.Moeller@uni-siegen.de}{2}
\addauthor{Juergen Gall}{gall@iai.uni-bonn.de}{1}

\addinstitution{
	University of Bonn
}
\addinstitution{
University of Siegen
}

\runninghead{Briq, Moeller, Gall}{CSPN}

\def\ie{\emph{i.e}\bmvaOneDot}

\DeclareMathOperator*{\argmin}{\mathrm{argmin}}
\DeclareMathOperator*{\argmax}{\mathrm{argmax}}
\DeclareMathOperator*{\minw}{\mathrm{min}}

\usepackage[]{algorithm2e}
\usepackage{amssymb}
\newcommand{\R}{\mathbb{R}}

\SetKwInput{KwInput}{Input}
\SetKwInput{KwOutput}{Output}
\SetKw{kwInit}{Initialize}
\SetKwRepeat{While}{while}

\begin{document}
	
	\maketitle
	
	\begin{abstract}
Weakly supervised semantic segmentation has been a subject of increased interest due to the scarcity of fully annotated images.
We introduce a new approach for solving weakly supervised semantic segmentation with deep Convolutional Neural Networks (CNNs). The method introduces a novel layer which applies simplex projection on the output of a neural network using area constraints of class objects. The proposed method is general and can be seamlessly integrated into any CNN architecture. Moreover, the projection layer allows strongly supervised models to be adapted to weakly supervised models effortlessly by substituting ground truth labels. Our experiments have shown that applying such an operation on the output of a CNN improves the accuracy of semantic segmentation in a weakly supervised setting with image-level labels.
		
	\end{abstract}
	
	\section{Introduction}
	\label{sec:intro}
	
The task of semantic image segmentation, which requires solving the problem of assigning a semantic class label to each pixel in a given image, is considered crucial in solving more complex tasks such as scene understanding. The most successful approaches for semantic image segmentation rely on convolutional neural networks (CNN). CNNs, however, require large training sets where each image is pixel-wise annotated. Since obtaining dense pixel annotations for training data is an expensive and tedious task, there has been a growing effort to learn CNNs using less supervision. In particular, a weakly supervised setting in which only image-level labels are available has been addressed in several works~\citep{papandreou2015weakly, Shimoda2016DistinctCS, kolesnikov2016seed, zhang2018decoupled, chaudhry2017discovering}. 

In order to guide the training process, a prior on the size of objects in an image is often used. For instance, \citet{chen2018deeplab} employ an expectation-maximization approach, which alternates between predicting the output of the network and estimating its parameters, which is refined by size priors. Instead of using a prior, \citet{pathak2015constrained} introduce constraints on the size of the objects. While constraints have the advantage of reducing the set of feasible solutions and are a more principled way to integrate prior knowledge about object sizes during training of a CNN, current state-of-the-art methods for weakly supervised semantic image segmentation do not rely on constraints. This is due to the way the approach in \cite{pathak2015constrained} enforces constraints on the output of a neural network. It proposes predicting a latent distribution for which violations of predefined linear constraints are penalized, and then optimizes the parameters of the CNN to follow the latent distribution as closely as possible by minimizing the Kullback-Leibler divergence. This means that an additional optimization step needs to be performed in each training iteration. 

In this work, we provide a more practical approach that integrates constraints directly as a network layer and can therefore be added to any convolutional neural network. We term the layer \emph{simplex projection layer} since it projects the output of the previous layer onto the simplex and enforces the network to satisfy the given constraints. Such an operation can be efficiently performed in linear time~\cite{duchi2008efficient}. 

To demonstrate the benefit of the simplex projection layer, we integrate the layer into a state-of-the-art approach for weakly supervised semantic image segmentation~\cite{kolesnikov2016seed}. In our experiments on the PASCAL VOC 2012 segmentation benchmark~\cite{everingham2015pascal}, we show that the layer substantially improves the segmentation accuracy of ~\cite{kolesnikov2016seed}. 
The source code is available at \url{https://github.com/briqr/CSPN}.	
	
	\section{Related Work}
	
Over the past few years, CNNs have made significant advances in improving the results in semantic segmentation, particularly in a strongly supervised setting~\citep{chen2018deeplab}. In such a setting, however, obtaining fully annotated data requires much effort and time and therefore limits their availability. To circumvent this limitation, several approaches that rely on weaker forms of supervision have been proposed. For instance, ~\citep{papandreou2015weakly} propose using bounding boxes in their training data in a semi-supervised setting. \citet{lin2016scribblesup} proposed using scribbles which mark the center of an object in a given image. ~\citet{chaudhry2017discovering} combine saliency and attention maps that form pseudo ground truth during training.

So far, the least expensive methods rely on image-level labels only. \citet{pinheiro2015image}, for example, reformulate the problem as multiple instance learning (MIL), such that an image is considered a positive bag if it contains at least one pixel of a certain class and negative otherwise. ~\citet{chen2018deeplab} employ an Expectation Maximization (EM) algorithm which alternates between predicting the pixel labels and optimizing the network parameters. ~\citet{Shimoda2016DistinctCS} employ a method based on back-propagation in which they compute the CNN derivatives with respect to the intermediate layers rather than the input images and then use subtraction to create class-specific saliency maps. \citet{pathakICLR15} propose a method that formulates MIL in a fully convolutional network, in which the loss is computed coarsely at maximum predictions in the heat map. ~\citet{saleh2016built} create a foreground/background mask by exploiting
the unit activations of some hidden layers in a fully convolutional CNN that had been pre-trained for object recognition.
~\citet{qi2016augmented} combine semantic segmentation and object localization modules in the CNN that provide augmented feedback to each other for error correction.~\citet{kim2017two} train two Fully Convolutional Networks (FCN) for image-level classification in two phases, where in the second phase the second FCN is trained with the highlighted regions in the heat maps from the first FCN being suppressed. Other methods rely on using external modules to generate cues such as localization and size information~\cite{kolesnikov2016seed, chaudhry2017discovering}. Such a module may either have been learned in a weakly supervised setting or trained on a different dataset in a more strongly supervised manner. 

\section{Convolutional Simplex Projection Network (CSPN)}

The task of semantic image segmentation solves the problem of assigning semantic class labels to each pixel in an image, \ie, for a given image $X$ of an arbitrary size ${m\times n}$ and a set of possible semantic labels denoted by $L$, the goal is to determine a label $k\in L$ for each pixel $x_{ij}\in X$, $i\in[m], j\in[n]$. Such a task can be solved by a CNN by estimating $Y=(y_{ij})$ with $y_{ij}=\argmax_{k\in L} (Q_k(X;\Theta))_{ij}$, which takes the image $X$ as input and predicts for each pixel $(i,j)$ the class label $y_{ij}$. $Q(X;\Theta)$ denotes the network output function and the variable $\Theta$ represents the network parameters being optimized during training. 

In the context of weakly supervised learning, pixel-wise labels are not provided during training. Instead, the network parameters $\Theta$ are learned based on a set of images, in which each training sample is a pair $(X, S)$, where $S\subseteq L\setminus \{k_0\}$ denotes the semantic classes present in the image $X$ and $k_0$ denotes the background label. In order to guide the training process, approaches for weakly supervised learning like \cite{chen2018deeplab} use size priors for semantic classes as additional weak supervision. Size priors were rephrased as constraints in the work of ~\cite{pathak2015constrained}. Their method uses a penalty function for violated linear constraints in order to estimate a latent distribution and optimizes the parameters of the CNN to follow the latent distribution as closely as possible. The approach alternates between updating the latent distribution using a dual formulation and updating the network parameters. In particular, the constraints are enforced during training only and cannot be guaranteed during inference. Furthermore, it is not straight forward to combine the approach with more advanced approaches for weakly supervised image segmentation, which outperform \cite{pathak2015constrained} by a large margin. In this work, we provide a more elegant approach that integrates constraints directly as a network layer and can therefore be added to any convolutional neural network.         

We first introduce the simplex projection layer in Section~\ref{sec:layer}. We then describe in Section~\ref{sec:CSPN} how such a layer can be applied for weakly supervised semantic segmentation using any CNN for image segmentation and finally describe an implementation, which adds the simplex projection layer to SEC~\cite{kolesnikov2016seed}. 

\subsection{Simplex Projection Layer}\label{sec:layer}
In general, projection onto the simplex is a minimization problem defined as:
\begin{equation}\label{eq:simplex_project}
\minw_w \frac{1}{2}||w-v||_2^2 \quad \textrm{s.t.} \sum_{i=1}^{n}w_i=\upsilon, \quad w_i\geq0,
\end{equation}
i.e. given a vector $v\in\R^n$ that does not satisfy a given equality constraint, it finds a new vector $w\in\R^n$ that satisfies the constraint and that is as close to $v$ as possible.
In the context of our problem, the goal is to train a network $Q(X;\Theta)$ that receives an image $X$ as input, and outputs a pixel-dense semantic segmentation $Y$ with $y_{ij}=\argmax_{k\in L} (Q_k(X;\Theta))_{ij}$. For integrating our approach into such a network, let us assume that the precise information about the size of the object $k\in L$ present in the image is $\upsilon_k$ pixels. In Section~\ref{sec:CSPN}, we describe how $\upsilon$ can be obtained but the approach is agnostic to how such an estimate may be obtained. 

The constraint that object $k$ should have size $\upsilon_k$ can be formalized by requiring the output $Q(X;\Theta)$ of the network for an object $k$ to be in the constraint set $C_Y$ defined as 
	\begin{equation}
	C_Y=\{Y | \sum_{i,j}y_{k_{i,j}}=\upsilon_k\},
	\end{equation}
	\ie only solutions where the number of pixels that are classified as class $k$ is equal to $\upsilon_k$ are valid. Since $\argmax$ is non-differentiable and the loss of the network is computed on $Q(X;\Theta)$, we instead approximate the constraint set by     
\begin{equation}
	C_Q=\{Q | \sum_{i,j}q_{k_{i,j}}=\upsilon_k\},
\end{equation}
\ie we require that the sum of the elements of the heat map of class $k$ be equal to the size $\upsilon_k$. 
	In order to enforce such constraints, we introduce the projection onto the set $C_Q$,
		\begin{equation}\label{eq:project}
		\pi_C(Q_k)=\argmin_{V_k\in C_Q}||Q_k-V_k||_F^2,
		\end{equation}
	as a layer into the network, where $Q_k$ is the output of the previous layer for class $k\in L$ and $||Q_k-V_k||_F^2=\sum_{i,j}(q_{k_{i,j}}-v_{k_{i,j}})^2$ denotes the squared Frobenius norm.   
	
	The projection onto the simplex $\pi_C(Q)$ can be efficiently solved by Algorithm~\ref{alg:simplexprojection}~\citep{duchi2008efficient}. In our context, the simplex projection layer takes as input a confidence map for class $k$ and a constraint value $\upsilon_k$ and outputs a refined map, which satisfies the constraint. In essence, the algorithm finds the elements of the heat map whose sum is used to calculate the difference from the true size. Once these elements are found, the difference is distributed evenly across the heat map. This solution is dictated by solving the Lagrangian formulation of the problem in  \eqref{eq:simplex_project}.
	The operations of the simplex projection layer are very inexpensive and the expected time complexity is linear, as such, integrating such a projection layer in a CNN does not increase its time complexity. Furthermore, the projection layer is not limited to a certain network architecture and can in principle be applied in any weakly supervised setting. The algorithm, however, requires prior knowledge of the size of an object in a given image, and given a weakly supervised setting such a number can only be estimated as we will discuss in the following section.

		\begin{algorithm}[t]
		\KwInput{A matrix $M\in \R^{m\times n}$ and a scalar $\upsilon$ > 0 }
		\kwInit {$U=[m]\times [n] \quad s=0 \quad \rho=0$}
		\SetKwProg{simplexprojection}{}{}{}
		\simplexprojection{}{
			\nl While $U\ne\emptyset$ \\
			\nl $\quad$ Pick $(g,h)\in U$ at random \\
			\nl $\quad$ Partition $U$: \\
			\nl	$\quad\quad$ $G=\{(i, j)\in U|M_{ij}\geq M_{g,h}\}$ \\
			\nl	$\quad\quad$ $L=\{(i, j)\in U|M_{ij}<M_{g,h}\}$ \;\
			\nl $\quad$ Calculate $\Delta\rho=|G| \;; \quad \Delta s = \sum_{(i,j)\in G}M_{i,j}$ \\
			\nl $\quad$ IF $(s+\Delta s)-(p+\Delta p)M_{g,h}<\upsilon$ \\
			\nl 	$\quad\quad$ $s=s+\Delta s \;; \quad  \rho=\rho+\Delta\rho \,;\quad U\leftarrow L $ \\
			\nl $\quad$ ELSE \\
			\nl $\quad\quad$ $U\leftarrow G \setminus \{(g,h)\}$ \\
			\nl $\quad$ ENDIF \\
			\nl $\quad$ SET $\theta=\frac{s-\upsilon}{\rho}$\\
			\nl OUTPUT $M_{ij}=max\{M_{i,j}-\theta,0\}$ \\
		} 
		\caption{Projection onto the simplex. The matrix $M$ represents a heat map containing the softmax probabilities for a class $k$ and $\upsilon$ is a scalar denoting the object size of class $k$. The algorithm has resemblance to the median-finding algorithm in which the idea is to reduce the time complexity by avoiding sorting the elements.}
		\label{alg:simplexprojection}
	\end{algorithm} 

\subsection{CSPN for Weakly Supervised Semantic Segmentation}\label{sec:CSPN}

\begin{figure}[t]
		\centering
		\includegraphics[width=12cm]{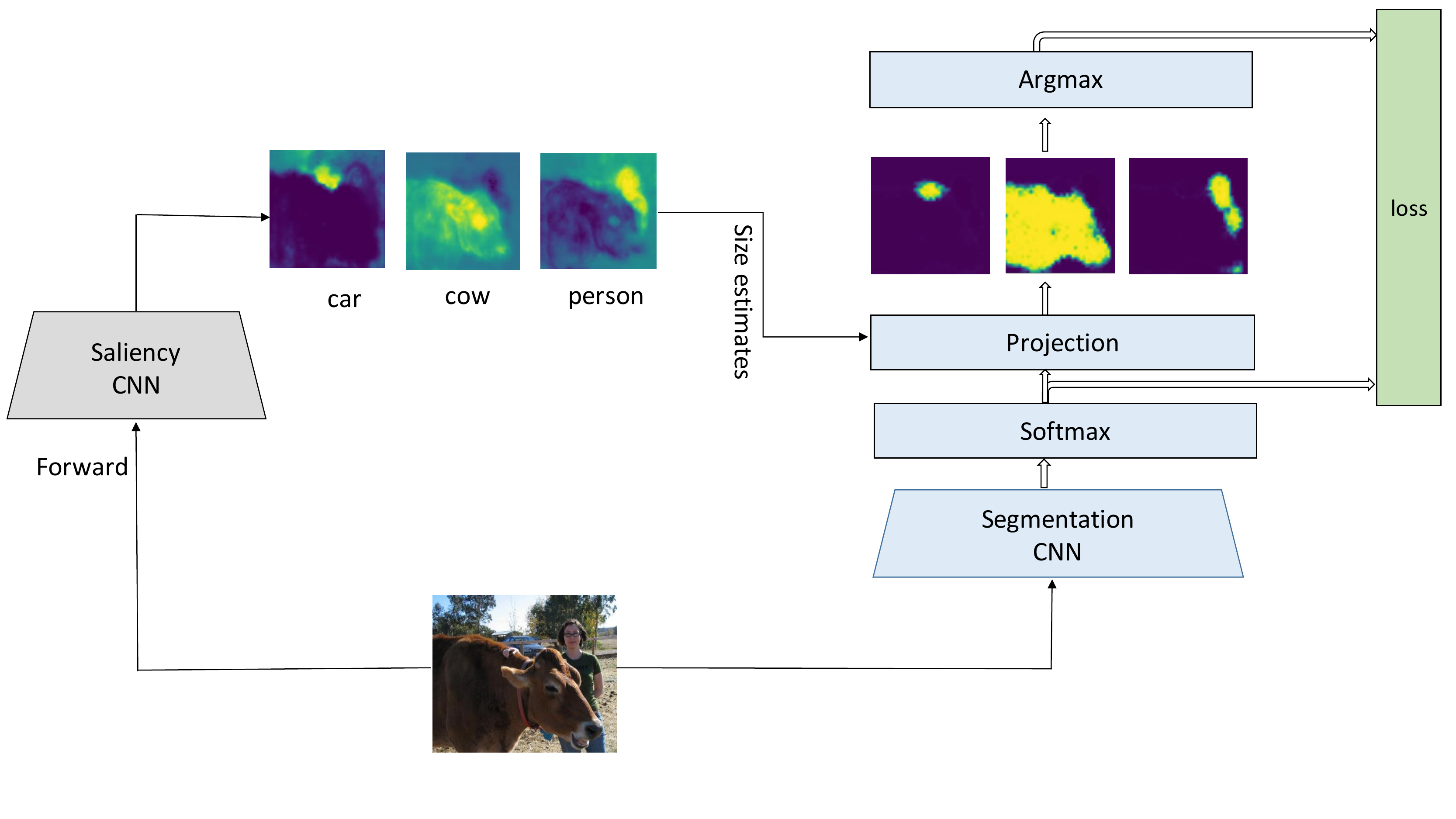}
		\caption{Illustration of the general approach with a projection layer in a weakly supervised setting. In our implementation, a saliency CNN provides saliency maps for each class in order to supply size estimates to the projection layer. These maps can be computed offline in advance. The projection layer outputs refined maps based on these estimates. The argmax operation is then applied to these heat maps thereby generating a mask that serves as ground truth according to which the loss function is computed. Note that for illustration purposes, the heat maps plotted after the projection layer have undergone a softmax operation, but this is not necessary in the network due to the argmax layer.
			}
		\label{fig:approach}
	\end{figure}

\begin{figure}[t]
	\centering
	\setlength\tabcolsep{5pt}
	\scalebox{0.81}{
		\begin{tabular}{llllll}
			\multicolumn{1}{c}{\includegraphics[scale=0.2]{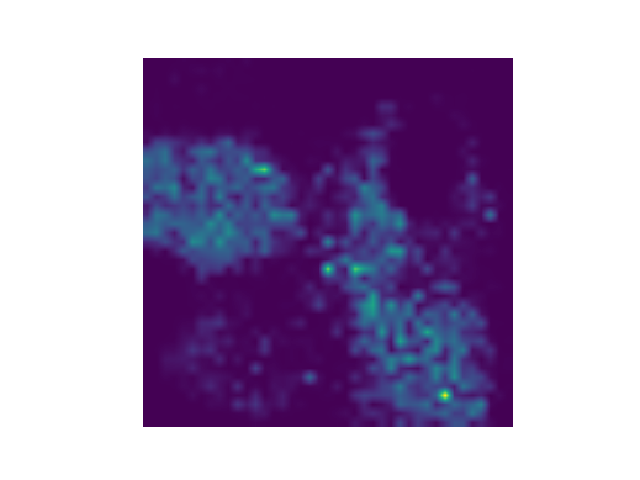}}&  
			\multicolumn{1}{c}{\includegraphics[scale=0.2]{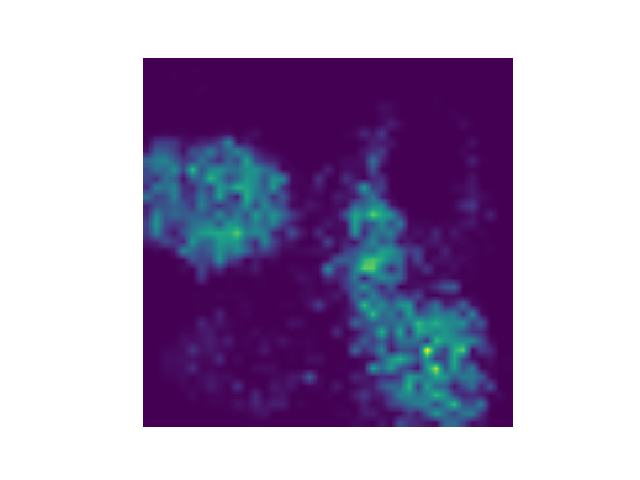}}&  
			\multicolumn{1}{c}{\includegraphics[scale=0.2]{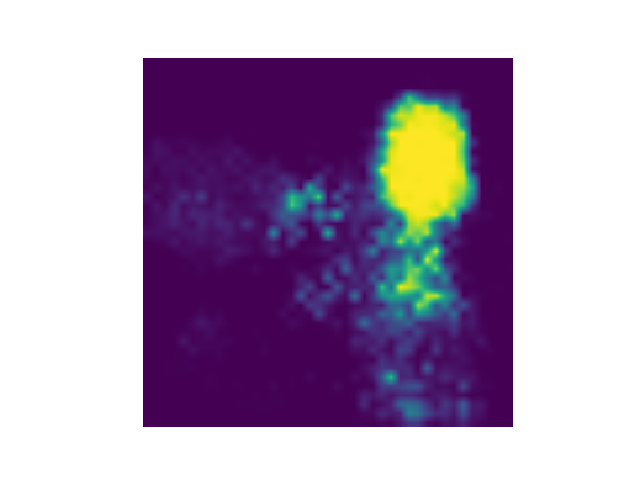}}&  \\
			\multicolumn{1}{c}{\includegraphics[scale=0.2]{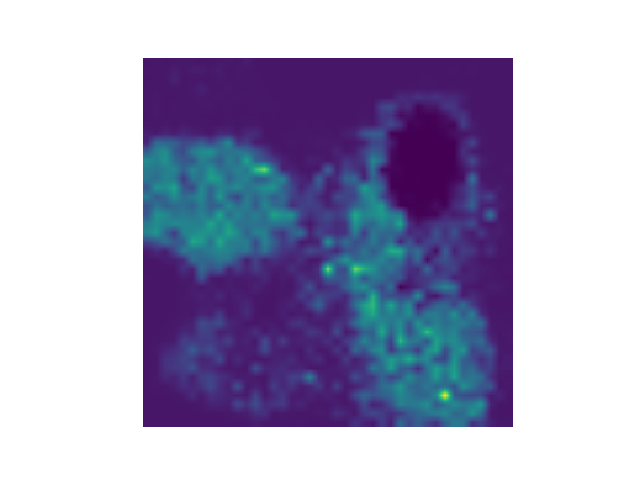}}&  
			\multicolumn{1}{c}{\includegraphics[scale=0.2]{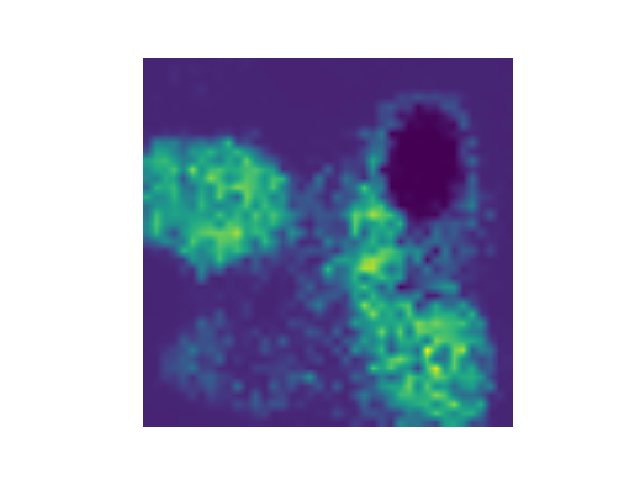}}&  
			\multicolumn{1}{c}{\includegraphics[scale=0.2]{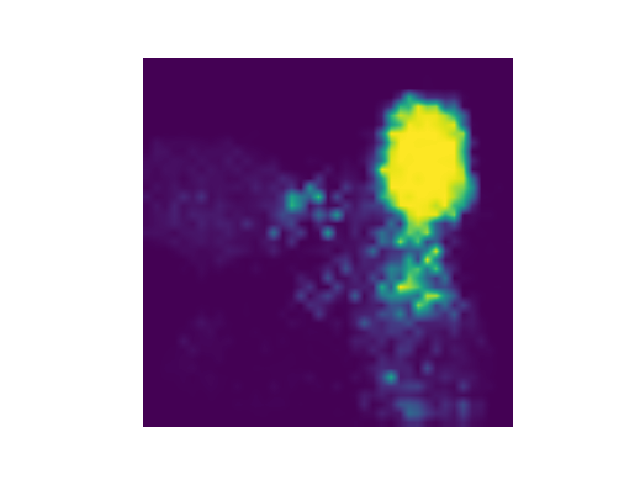}} \\ 
			
			\multicolumn{1}{c}{car}&  \multicolumn{1}{c}{cow}&  \multicolumn{1}{c}{person}&  \multicolumn{1}{c}{}&  
		\end{tabular}}
		\caption{Visualisation of the effect of the projection layer at an early stage of training. The simplex projection is applied to the heat maps of the last convolution layer in the CNN, namely fc8. The first row is the heat map of fc8, and the second row is the projected heat map. As can be seen, the projection layer enhances the heat maps of the CNN such that the detected objects become more prominent in the projected heat map.}
		\label{fig:heatmaps}
	\end{figure}

Figure~\ref{fig:approach} gives an overview of how the simplex projection layer can be applied in a weakly supervised setting, in which only image-level labels are available. In order to enforce some given constraints at the last layer, we introduce a softmax layer after the last fully convolutional layer in the network, which performs:
		\begin{equation}
		\sigma(Q_k(x_{ij})) = \frac{e^{Q_k(x_{ij})}}{\sum_{l=1}^{L}e^{Q_l(x_{ij})}}
		\end{equation}
	for every class $k\in L$ and pixel in the output image.
	The softmax layer is followed by the simplex projection layer which takes as input the probability values from the softmax layer, the size of each object present in the image, and outputs the projection onto the constraint set $C_Q$. Figure~\ref{fig:heatmaps} visualises the effect of the projection layer.   
	
	An argmax operation is applied to the output of the simplex projection layer to create the target label for the loss layer

	\begin{equation}\label{eq:max}
	\hat{Y} = \argmax_{k\in L}\pi_C(Q_k).
	\end{equation}

	As a loss function, we use the standard cross-entropy loss given by:
	\begin{equation}
	L_{project}=-\sum_{ij}\sum_{k=1}^{L}\hat{y}_{k_{ij}} \log \left((Q_k(X;\Theta\left)\right)_{ij} \right ) 
	\label{eq:entropyloss}
	\end{equation}
	where $\Theta$ is the set of the CNN parameters, $\hat{y}_{k_{ij}}$ is one if the pixel $x_{ij}$ is assigned to class $k$ after the projection \eqref{eq:max} and zero otherwise, and $\left(Q_k(X;\Theta)\right)_{ij}$ denotes the predicted class probability at pixel $x_{ij}$ before the simplex projection layer. 

	 Since we are tackling semantic segmentation in a weakly supervised setting, the object size $\upsilon_k$ is not available at hand except if the class $k$ is not present in the image in which case $\upsilon_k=0$. To obtain an estimate of $\upsilon_k$, we use the approach in ~\cite{Shimoda2016DistinctCS} that estimates class-specific saliency maps.
	 
While negative values in the maps mark non-salient regions, positive values indicate potential salient regions. To obtain $\upsilon_k$ we therefore count the number of pixels that have a score larger or equal to $\tau=\frac{1}{8}$. To avoid counting a pixel twice for different classes, we only count the pixels for class $k$ with the highest saliency score.    
In our experiments, we show that the saliency maps estimated by~\cite{Shimoda2016DistinctCS} are not very accurate, but they provide a reasonable estimate for $\upsilon_k$. It is noteworthy that the approach~\cite{Shimoda2016DistinctCS} is trained in the same weakly supervised setting and does not require any additional supervision or additional training data. Even though we have used saliency maps to obtain these estimates, the approach is independent of what estimator may be used.

\subsection{Simplex Projection Layer in SEC}\label{sec:SEC} 

While the previous section described how the simplex projection layer can be added to any CNN for weakly supervised image segmentation, we now describe its integration into the state-of-the-art SEC~\cite{kolesnikov2016seed} approach. The loss of SEC consists of three terms, namely seed, expand and constrain (SEC), which encourage the network to meet localization cues for an object, predict its right extent and to meet its precise boundaries respectively. The overall loss in SEC is given by:
		\begin{equation}
		L=\sum_{(X,S)}\left( L_{seed}(Q(X;\Theta),S)+L_{expand}(Q(X;\Theta),S)+L_{constrain}(Q(X;\Theta),S) \right).
		\label{eq:sec_loss}
		\end{equation}
		For a detailed description of the tree terms, we refer to~\cite{kolesnikov2016seed}.

	In our work, we replace the expand loss with our loss based on the simplex projection~\eqref{eq:entropyloss}: 
\begin{equation}
			L=\sum_{(X,S)}\left( L_{seed}(Q(X;\Theta),S)+L_{project}(Q(X;\Theta),S)+L_{constrain}(Q(X;\Theta),S) \right).
			\label{eq:spc_loss}
\end{equation}
To compute $L_{project}$, we append the simplex projection layer to the last fully convolutional layer fc8 in the CNN. 

\begin{figure}[t]
		\centering
		\includegraphics[width=8cm]{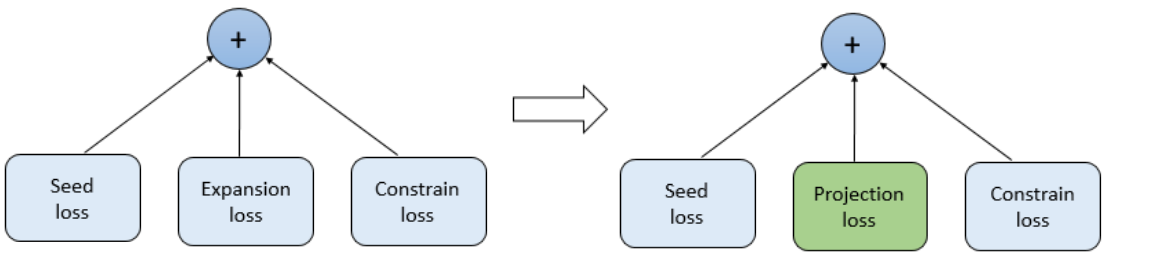}
		\caption{This figure shows how the loss function has been adapted from SEC to our proposed approach. The loss of SEC consists of three loss functions (left). In our approach (right), we omit the expansion loss and add the projection loss.}
		\label{SC_projection}
	\end{figure}

In principle, $L_{expand}$ and $L_{project}$ can be combined but the terms are redundant and there is no benefit of adding both terms. While~\cite{kolesnikov2016seed} uses the expand loss in order to expand a detected object to a reasonable size, the projection loss projects the solution onto the constraint set which ensures that each object has a reasonable size. $L_{project}$ can therefore be considered as a more strict loss function than $L_{expand}$. The difference between the two loss functions~\eqref{eq:sec_loss} and~\eqref{eq:spc_loss} is illustrated in Figure~\ref{SC_projection}.
The optimal performance was achieved after about $4-5$ epochs. This fast convergence can be attributed to the fact that the space of feasible solutions is reduced to only those that are within the constraint set.

		\begin{table}[t]
		
		\begin{tabular}{l*{6}{c}r}
			\hline
			
			Method              & mIoU & Additional supervision \\
		\hline
	
		DCSP-VGG16 \citep{chaudhry2017discovering} & 58.6 & supervised saliency& \\
		DCSP-ResNet-101 \citep{chaudhry2017discovering} & 60.8 & supervised saliency& \\
		AE w/o PSL \citep{wei2017object} &50.9 & supervised saliency &\\
		AE-PSL \citep{wei2017object} &55.0 & supervised saliency &\\
		\hline\hline
			
			MIL-FCN \citep{pathakICLR15} & 25.7 & \hspace{44.4mm} \\
			EM-Adapt \citep{papandreou2015weakly}  & 38.2 & \\
			CCNN \citep{pathak2015constrained}  & 35.3 & \\
			BFBP \citep{saleh2016built} & 46.6 & \\
			DCSM \citep{Shimoda2016DistinctCS} & 44.1 & \\
			SEC \citep{kolesnikov2016seed} &50.7 & \\
			AF-SS \citep{qi2016augmented} & 52.6 & \\
			Two-phase \citep{kim2017two} & 53.1 & \\
			\citet{roy2017combining} & 52.8 & \\
			DCNA-VGG16 \citet{zhang2018decoupled} &55.4&\\
			DCNA-ResNet-101 \citet{zhang2018decoupled} &58.2&\\
			CSPN(ours) &54.5&\\ 						
			\hline
		\end{tabular}
		\caption{Results on the validation set.} \label{tab:val_evaluation}
		\end{table}

	\begin{table}[t]
		\begin{tabular}{l*{6}{c}}
			\hline
			Method              & mIoU  \\
		\hline
			MIL-FCN \citep{pathakICLR15} & 24.9 & \hspace{44.4mm} \\
			EM-Adapt \citep{papandreou2015weakly}  & 39.6 & \\
			CCNN \citep{pathak2015constrained}  & 35.3 & \\
			BFBP \citep{saleh2016built} & 48.0 & \\
			DCSM \citep{Shimoda2016DistinctCS} & 45.1 & \\
			SEC \citep{kolesnikov2016seed} &51.7 & \\
			AF-SS \citep{qi2016augmented} & 52.7 & \\
			Two-phase \citep{kim2017two} & 53.8 & \\
			\citet{roy2017combining} & 55.5 & \\
				DCNA-VGG16 \citet{zhang2018decoupled} &56.4&\\
				DCNA-ResNet-101 \citet{zhang2018decoupled} &60.1&\\
			CSPN(ours) &55.5&\\ 						
			\hline
		\end{tabular}
		  \caption{Results on the test set.} \label{tab:test_evaluation}
		\end{table}

	\section{Evaluation}
	The proposed method has been trained and evaluated on the PASCAL VOC 2012 segmentation benchmark~\citep{everingham2015pascal}, which contains 10,582  training, 1449 validation and 1456 test images. The implementation is based on the Deeplab architecture~\citep{chen2018deeplab}, refined by~\citet {kolesnikov2016seed} to integrate the SEC loss, and the Caffe deep learning framework ~\citep{jia2014caffe}. Conforming with common practice, the results are refined by a CRF in a post-processing step.
	
	\subsection{Comparison with State-of-the-Art}
	
	Tables \ref{tab:val_evaluation} and \ref{tab:test_evaluation} compare the accuracy of our approach (CSPN) with the state-of-the-art for weakly supervised image segmentation for the validation and test sets respectively. The numbers of the other approaches are taken from~\cite{zhang2018decoupled}. On the validation set, CSPN outperforms DCSM~\citep{Shimoda2016DistinctCS} and SEC \citep{kolesnikov2016seed} by +10.4\% and +3.8\%, respectively, and on the test set by +10.4\% and +3.8\%. This shows that the simplex projection layer substantially improves the accuracy of SEC. 
	
	Our approach also outperforms by a large margin the expectation-maximization approach proposed by  \citet{papandreou2015weakly} and the constrained convolutional neural network (CCNN) \citep{pathak2015constrained}. Although \citet{pathak2015constrained} showed in their work that CCNN improves \cite{papandreou2015weakly} slightly by +1.5\% on the validation set, the accuracy of CCNN is much lower compared to \cite{papandreou2015weakly} in Table~\ref{tab:val_evaluation}. This is due to the fact that CCNN builds on a preliminary version of~\cite{papandreou2015weakly}. In order to evaluate how much improvement our approach would attain when integrated in \cite{papandreou2015weakly}, we added our simplex projection layer to the public available implementation of \cite{papandreou2015weakly}. In this case, the accuracy is improved by +6\%, which is clearly higher than the improvement which was reported in \cite{pathak2015constrained}. Since, however, both \cite{papandreou2015weakly} and \cite{pathak2015constrained} are inferior to the state-of-the-art in weakly supervised image segmentation, the reported numbers of CSPN build on the integration of the simplex projection layer in SEC as discussed in Section~\ref{sec:SEC}.           
	
	When we compare our approach with the state-of-the-art approaches~\cite{kim2017two,roy2017combining} that use the same amount of supervision, we see that our approach is slightly better or on par with them. Only the very recent work of DCNA \cite{zhang2018decoupled} outperforms our approach. Note that our model builds on SEC, which uses VGG16. We expect further improvements if ResNet-101 is used instead of VGG16. There are also approaches like DCSP~\cite{chaudhry2017discovering} which achieve a higher accuracy, but these approaches use additional supervision for training a saliency model. 

	\begin{table}[t]
	\centering
		\begin{tabular}{lccc|c}
			\hline
		Loss	& SEC+P & SC+P (CSPN) & S+P & SEC \\
		\hline
		mIoU & 51.0 & 54.5 & 44.6 & 50.7\\
			\hline
		\end{tabular}
		  \caption{Combining the projection loss (P) with additional loss functions from SEC.} \label{tab:SEC_loss}
		\end{table}

\subsection{Ablation Studies}

 In Section~\ref{sec:SEC}, we replace the expansion loss of SEC by the projection loss. In Table~\ref{tab:SEC_loss}, we report the accuracy if the expansion loss is added in addition or the constrain loss is omitted. Note that the seed loss is crucial for SEC and cannot be omitted. Adding the expansion loss to \eqref{eq:spc_loss} decreases the accuracy. Since both the expansion loss and projection loss ensure that the segmented objects have a reasonable size, adding both of them has the undesired effect of increasing the impact of the size compared to the seed and constrain loss. If the constrain loss is omitted, the accuracy also drops. This is expected since the simplex projection does not take object boundaries into account, which the constrain loss does using a CRF model based on the CNN output and image pixels. As such, the projection loss cannot compensate for the constrain loss.

\begin{figure}[t]
			\centering
			\includegraphics[width=8cm]{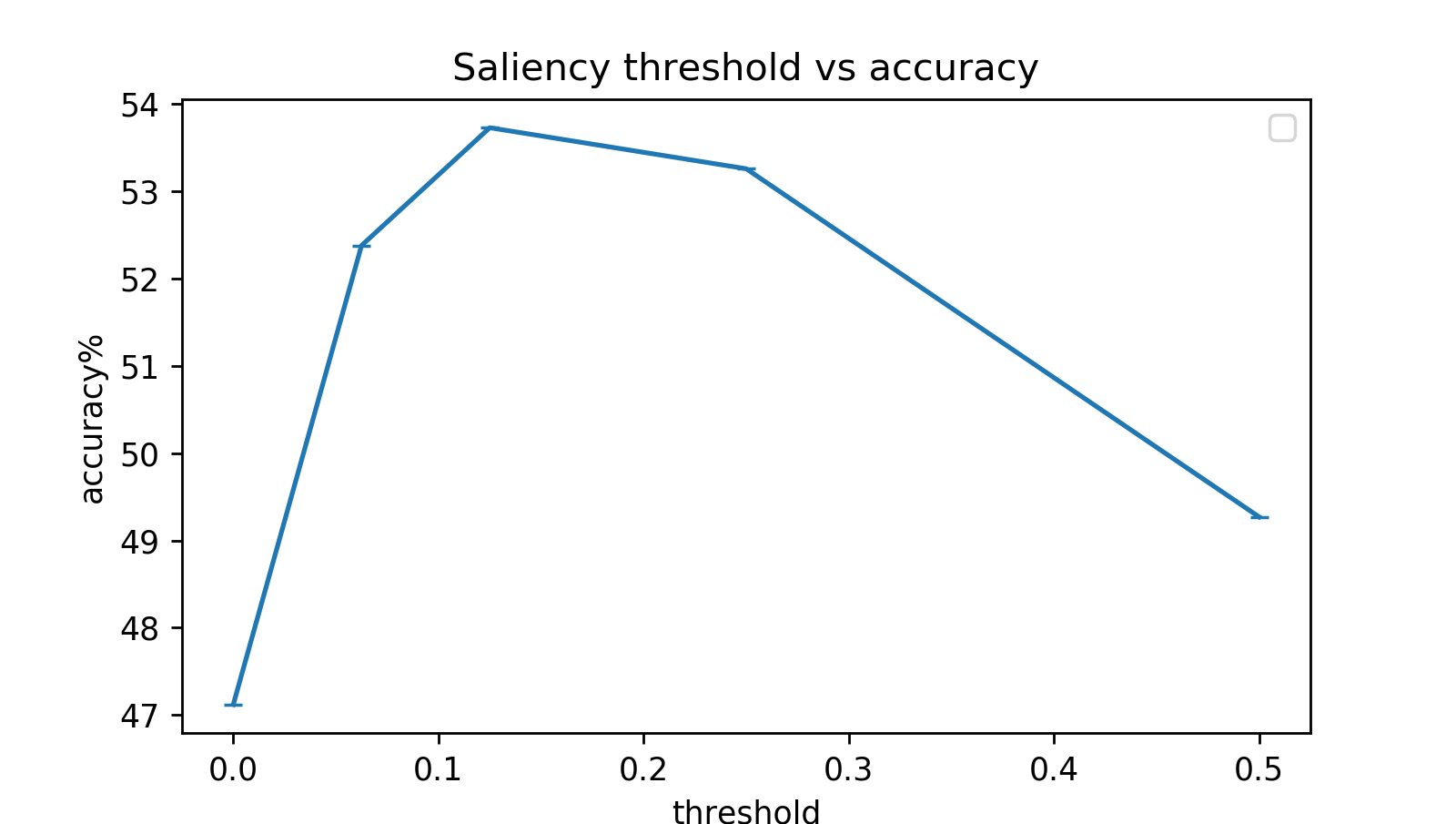}
			\caption{This plot examines the effect of the threshold value on the class-specific saliency. Values between $1/16$ and $1/4$ yield the most reliable values $\upsilon_k$ to be used in the projection.}
			\label{fig:threshold_vs_accuracy}
		\end{figure} 

In order to show that the merit of the method is due to the projection layer rather than the mere fusion of \citep{kolesnikov2016seed} and \citep{Shimoda2016DistinctCS}, we used the saliency of \citep{Shimoda2016DistinctCS} as the plain ground truth and obtained $51.7$, which only slightly improves over the $50.7$ of SEC, compared to $54.5$ when applying the projection.

As described in Section~\ref{sec:CSPN}, we use the threshold $\tau=1/8$ to obtain $\upsilon_k$ from the class-specific saliency maps~\cite{Shimoda2016DistinctCS}. The impact of the threshold is plotted in Figure~\ref{fig:threshold_vs_accuracy}. Note that the threshold has to be larger than zero since negative values or values around zero mark non-salient regions~\cite{Shimoda2016DistinctCS}.  The plot shows that a threshold between $1/16$ and $1/4$ works well. If the threshold is too high, \ie $1/2$ or larger, the size of the objects is underestimated and the accuracy decreases.

As illustrated in Figure~\ref{alg:simplexprojection}, we perform an argmax operation after the simplex projection layer. As an alternative, one could also add a softmax layer after the projection layer to obtain target class probabilities and use them for the loss function instead of $\hat{y}_{k_{ij}}$~\eqref{eq:entropyloss}. If we use softmax instead of argmax, the accuracy falls from 54.5\% to 50.2\%. 

\begin{figure}[t]
	\centering
	\setlength\tabcolsep{5pt}
	\scalebox{0.81}{
		\begin{tabular}{llllll}
			\multicolumn{1}{c}{\includegraphics[scale=0.13]{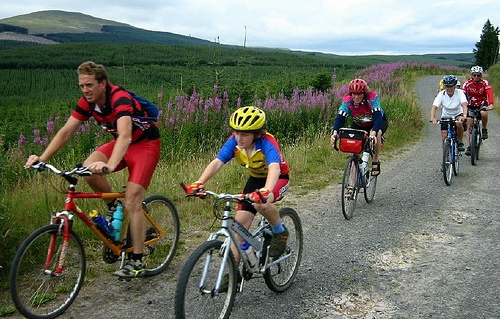}}&  
			\multicolumn{1}{c}{\includegraphics[scale=0.13]{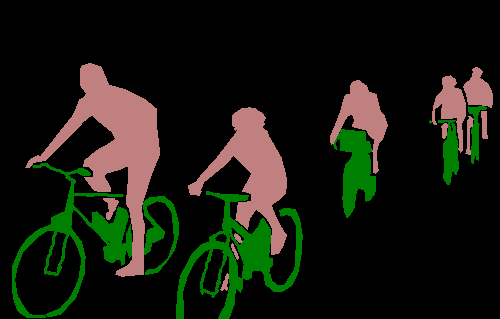}}&  
			\multicolumn{1}{c}{\includegraphics[scale=0.13]{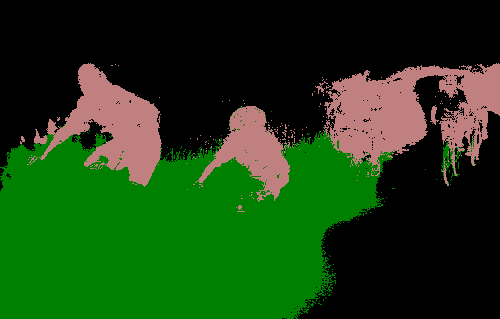}}&  
			
			\multicolumn{1}{c}{\includegraphics[scale=0.13]{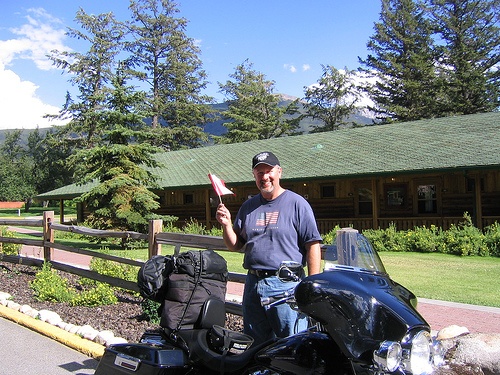}}&  
			\multicolumn{1}{c}{\includegraphics[scale=0.13]{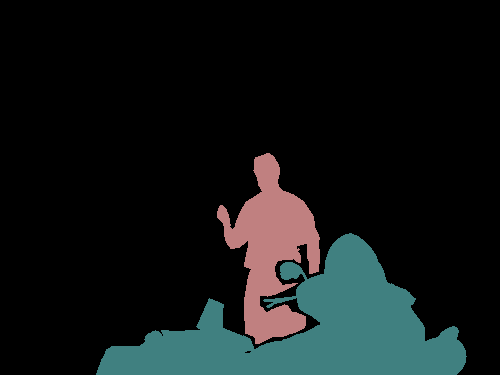}}&  
			\multicolumn{1}{c}{\includegraphics[scale=0.13]{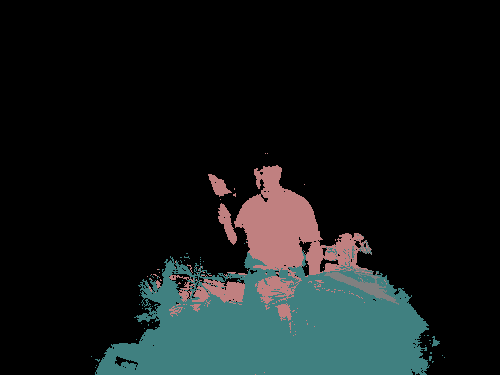}} \\ 
			
			\multicolumn{1}{c}{\includegraphics[scale=0.13]{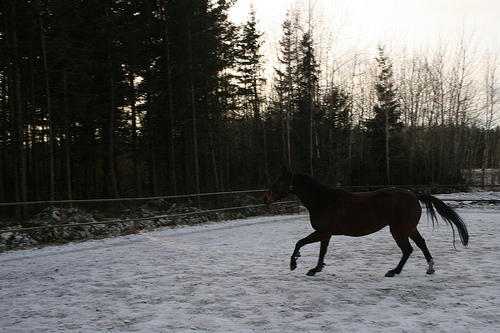}}&  
			\multicolumn{1}{c}{\includegraphics[scale=0.13]{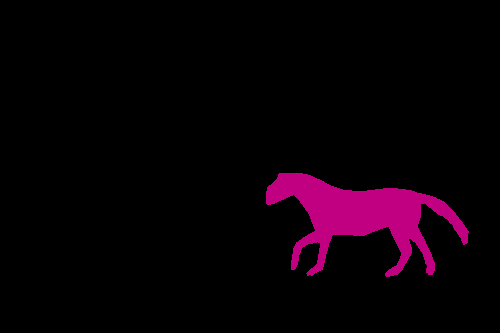}}&  
			\multicolumn{1}{c}{\includegraphics[scale=0.13]{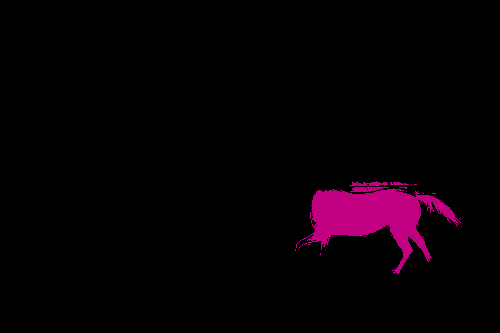}}&  
			
			\multicolumn{1}{c}{\includegraphics[scale=0.13]{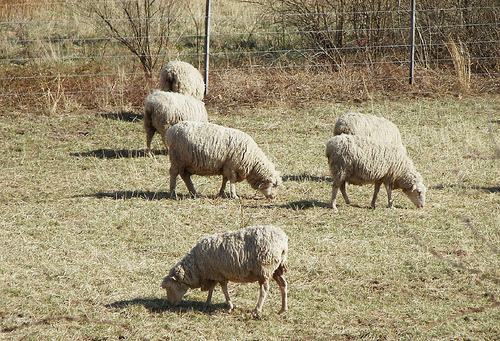}}&  
			\multicolumn{1}{c}{\includegraphics[scale=0.13]{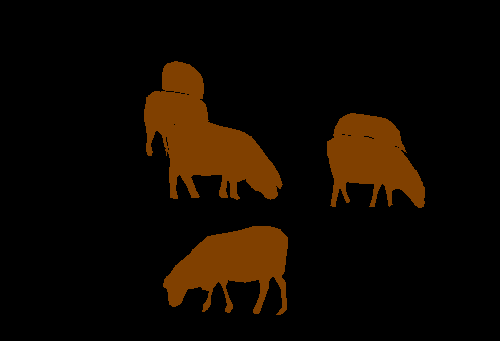}}&  
			\multicolumn{1}{c}{\includegraphics[scale=0.13]{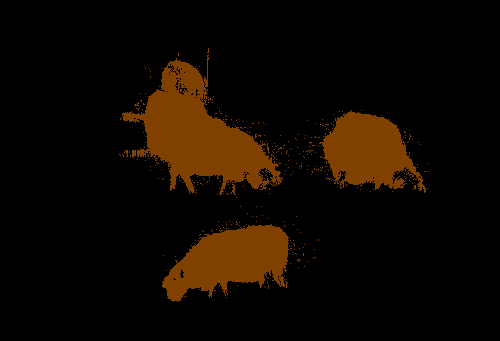}}  \\

			\multicolumn{1}{c}{\includegraphics[scale=0.13]{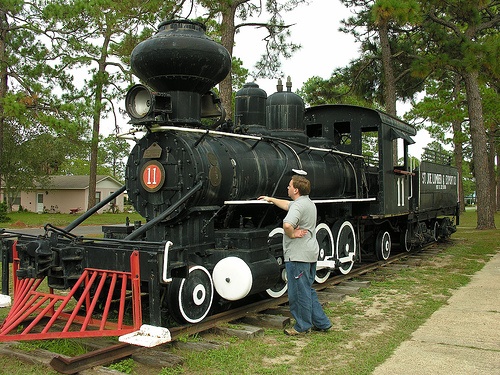}}&  
			\multicolumn{1}{c}{\includegraphics[scale=0.13]{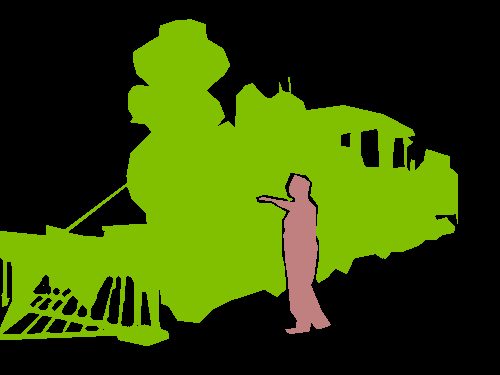}}&  
			\multicolumn{1}{c}{\includegraphics[scale=0.13]{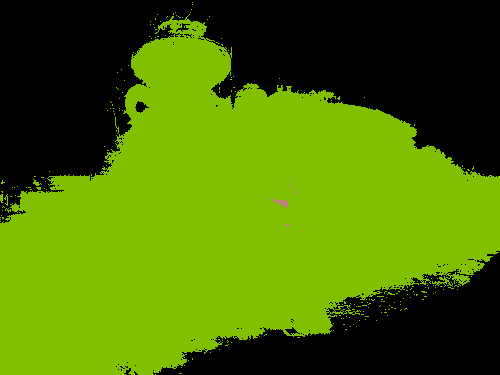}}&  
			
			\multicolumn{1}{c}{\includegraphics[scale=0.13]{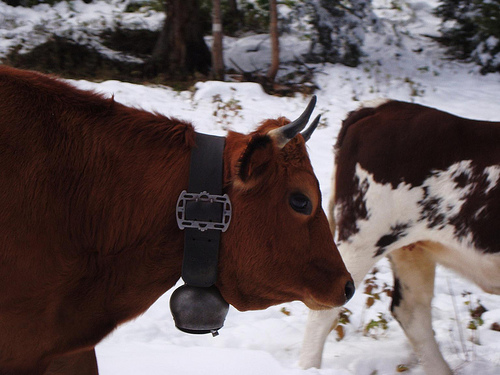}}&  
			\multicolumn{1}{c}{\includegraphics[scale=0.13]{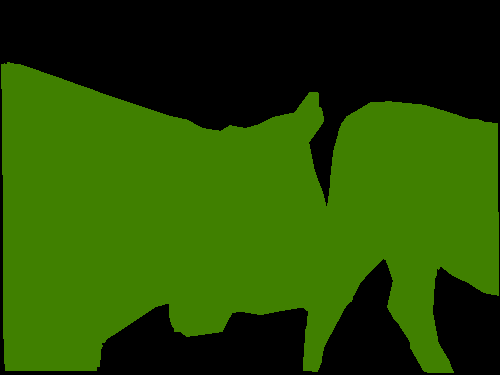}}&  
			\multicolumn{1}{c}{\includegraphics[scale=0.13]{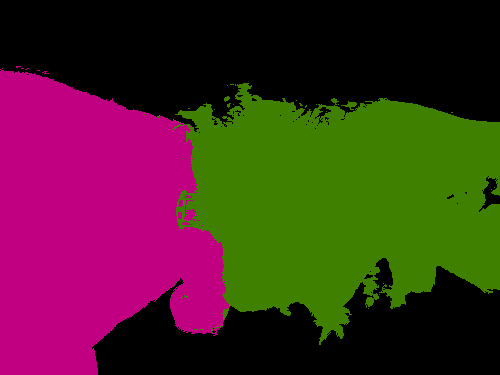}} \\  
			
			\multicolumn{1}{c}{Image}&  \multicolumn{1}{c}{Ground Truth}&  \multicolumn{1}{c}{Our}&  \multicolumn{1}{c}{Image}&  \multicolumn{1}{c}{Ground Truth}&  \multicolumn{1}{c}{Our}
		\end{tabular}}
		\caption{Above are some results obtained from evaluating our approach on the PASCAL VOC 2012 validation set, including two failure cases in the last row.}
		\label{fig:segmentationExamples}
	\end{figure}

A few qualitative results including failures cases are shown in Figure~\ref{fig:segmentationExamples}.

\section{Conclusions}
In this work, we have proposed the simplex projection layer which projects the output of a previous layer to the simplex. We have demonstrated the advantage of such layer for weakly supervised semantic segmentation where the projection layer is used to enforce constraints on the size of the objects. We integrated the layer in a state-of-the-art approach for weakly supervised segmentation and improved the segmentation accuracy substantially. As part of future work, we will investigate how the approach could be used for other tasks where the set of feasible solutions of a network can be reduced by constraints.      

\section*{Acknowledgements}
The work has been financially supported by the ERC Starting Grant ARCA (677650).	
	
	\bibliography{egbib}
\end{document}